\documentclass[journal]{IEEEtran}

\usepackage{cite}
\usepackage{amsmath,amssymb,amsfonts}
\usepackage{algorithmic}
\usepackage{colortbl}
\usepackage{graphicx}
\usepackage{textcomp}
\usepackage{framed,multirow}
\usepackage{mwe}
\usepackage{float}
\usepackage{comment}
\usepackage{xcolor}
\usepackage{multirow}
\usepackage{amsmath}
\usepackage{url}
\usepackage{hyperref}
\usepackage{adjustbox}
\usepackage{multirow}
\usepackage{breqn}
\usepackage{enumitem}
\usepackage{mathtools}
\usepackage{subcaption}

\definecolor{newcolor}{rgb}{.8,.349,.1}

\ifCLASSINFOpdf
\else
\fi

\begin{document}
 
\title{Eating Habits Discovery in Egocentric Photo-streams}

\author{Estefania~Talavera,
        Andreea~Glavan*\thanks{*Both authors contributed equally to this study},
        Alina~Matei*,
        and~Petia~Radeva, {\em IAPR Fellow}
\thanks{A. Glavan, A. Matei, and E. Talavera are with the Bernoulli Institute, University of Groningen.}
\thanks{P. Radeva is with the University of Barcelona and Computer Vision Center, Spain.}
\thanks{Manuscript received 2020; revised 2020.}}

\markboth{Journal of \LaTeX\ Class Files,~Vol.~14, No.~8, August~2015}%
{Shell \MakeLowercase{\textit{et al.}}: Eating Habits Discovery From Egocentric Photo-streams}

\maketitle


\begin{abstract}
Eating habits are learned throughout the early stages of our lives. However, it is not easy to be aware of how our food-related routine affects our healthy living. 
In this work, we address the unsupervised discovery of nutritional habits from egocentric photo-streams. We build a food-related behavioural pattern discovery model, which discloses nutritional routines from the activities performed throughout the days. To do so, we rely on Dynamic-Time-Warping for the evaluation of similarity among the collected days. Within this framework, we present  a simple, but robust and fast novel classification pipeline that outperforms the state-of-the-art on food-related image classification with a weighted accuracy and F-score of 70\% and 63\%, respectively.  Later, 
we identify days composed of nutritional activities that do not describe the habits of the person as anomalies in the daily life of the user with the Isolation Forest method. Furthermore, we show an application for the identification of food-related scenes when the camera wearer eats in isolation. 
Results have shown the good performance of the proposed model and its relevance to visualize the nutritional habits of individuals.  
\end{abstract}

\begin{IEEEkeywords}
Egocentric vision, Nutrition, Behaviour, Pattern recognition, Lifelogging.
\end{IEEEkeywords}
 
\IEEEpeerreviewmaketitle

\section{Introduction}

\IEEEPARstart{N}{utrition} 
has a significant influence on everyone's daily routine. As an example, studies have shown that American people spend on average 2.5h a day eating or drinking, out of which 78 minutes eating or drinking while doing other primary activities, such as driving, preparing meals or working \cite{food}. \textit{What} people eat has been regarded so far as the main factor impacting people's food behaviour. However, recent studies have shown that \textit{how} and \textit{where} people eat also play an important role \cite{wherepeopleeat}. Our hypothesis is that by gathering insight into the context of food-related activities, people can improve their eating habits with the view of leading a healthier lifestyle. Eating habits have a direct impact on one's health: diseases like diabetes, obesity, cardiovascular conditions, cancer and even mental illnesses are closely related to \textit{what} people eat \cite{cancer}\cite{obesity}\cite{mental}. Moreover, research has shown that loneliness has links to eating disorders \cite{lonely}, while television advertisements that promote calorie-dense foods and snacks have been shown to trigger mindless eating or snacking \cite{snacks}. Nowadays, it is common to entail isolated eating with eating alone in front of a television or in front of a computer. 

Food balance estimation has been studied from images intentionally collected by the user \cite{aizawa2013food}. 
The study of behaviour in our society has been addressed from crowds \cite{li2018deep} and individuals \cite{caip}. Information related to nutritional habits can aid nutritionists to achieve a better understanding of their patients' eating habits. Hence, such tools would enable them to provide to their patients with personalized advice regarding their lifestyle. 

\begin{figure}[h!] 
\centering
\begin{subfigure}[b]{0.47\textwidth}
   \includegraphics[width=1\linewidth]{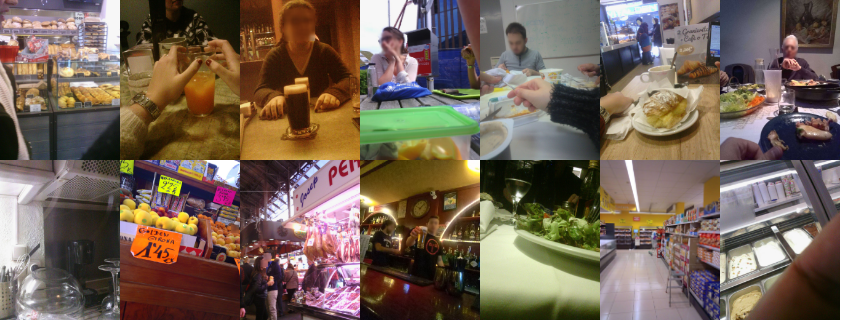}
   \caption{}
\end{subfigure}
\hfill 
\begin{subfigure}[b]{0.47\textwidth}
   \includegraphics[width=1\linewidth]{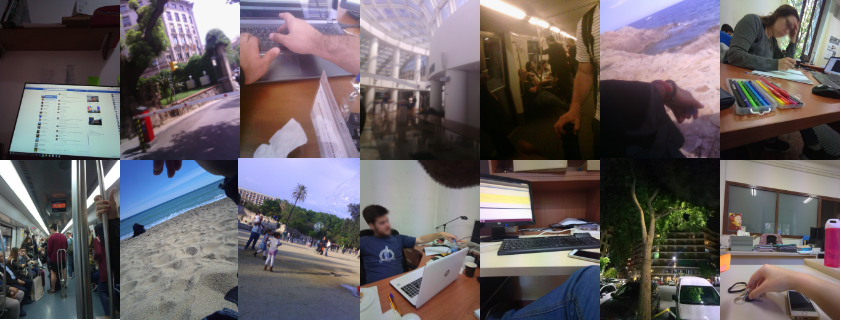}
   \caption{}
\end{subfigure}
\caption{Examples of egocentric images describing daily activities; we distinguish between images depicting (a) food-related and (b) non-food related activities and scenes.}
\label{fig:examples_mozaic}
\end{figure}

More specifically, \textit{lifelogging} \cite{lifelogging,lifelogging1} is an emerging mainstream activity in which day-to-day life activities are tracked using various types of wearable sensors. Egocentric photo-streams are collected by portable cameras worn as a necklace. These images provide a first-person perspective of the activities conducted by the camera's wearer. Some examples of sampled images from egocentric photo-streams can be seen in Fig. \ref{fig:examples_mozaic} below. The data is extremely meaningful for inferring and extracting patterns\cite{patterns,varini2017personalized} related to human behaviour. However, there is a lack of automated tools that can process egocentric photo-streams with the aim of providing insight into the context of food intake and nutritional behaviour. In this paper, we study the nutritional tendencies exhibited by people in their daily lives through the analysis of egocentric photo-streams.

For automatic analysis of nutritional habits, it is critical to be able to automatically process unseen collections of images. Therefore, the first step of our proposed pipeline is the classification of images into food or non-food related scenes. Later, we study the pattern of occurrence of the food-related images to find nutritional habits. To this end, first, a binary classifier filters food vs. non-food images. Then, images that receive the food label are further passed through the rest of the pipeline and labelled according to the 15 available food scene classes. Later, our proposed algorithm for nutritional habits discovery analyses the given labels using Dynamic Time Warping \cite{dtw}. This method allows us to measure the similarity among the photo-streams that describe the captured days. 

Routines differ per person and therefore cannot be detected by a supervised classifier, but should be discovered. Our hypothesis is that routine days should have similar appearance taking into account that events can be displaces in time (breakfast could occur in different hours in the morning). Hence, once defined the distance between the days, the routine will be discovered as a cluster of similar days. While non-routine  days will represent the outliers in this space. Therefore, we propose to analyse the collected data with the Isolation Forest anomaly detection method. This clustering method allows us to differentiate between Routine and Non-Routine days in an unsupervised way, discovering the different nutritional habits that compose them by identifying routine outliers. Furthermore, we study detected objects in food-related images in order to seek and quantify moments where the camera wearer eats in isolation.

The contributions of this work are three fold:
\begin{itemize}
\item We introduce the first automated tool for nutritional routine discovery from unseen egocentric photo-streams. Our model is capable of  detecting several routines in the life of the camera wearer in an unsupervised manner.

\item We present a new and simple method for the classification of images of given unseen egocentric photo-streams into non-food and food-related classes. 

\item We design an extension of our model for the detection of individual's isolation while eating.
    
\item We extend the EgoFoodPlaces dataset proposed by \cite{hierarchical} with 34084 new images describing non-food related scenes. Thus, our food-related scene classification algorithm is tested over a total of 68136 images.
\end{itemize}

This paper is structured as follows. In Section \ref{sec:related_works}, we present relevant works for the subject at hand. In Section \ref{sec:methodology}, we describe our proposed model and present the achieved results in Section \ref{Section4:results}. Finally, we draw conclusions and future work in Section \ref{Section6:Discussion}.

\section{Related works}
\label{sec:related_works}
The understanding of the behavior of people through the analysis of their collected egocentric photo-streams is a fairly new field. More specifically, there is little research available on the analysis of nutritional-related scenes in egocentric vision. 
The classification of food-related scenes from egocentric images was first presented in \cite{hierarchical}. The authors developed a hierarchical classification model and introduced a food-scene taxonomy as well as a new egocentric dataset, called \textit{EgoFoodPlaces}. The dataset is split semantically in meta-classes corresponding to nutrition-related activities (i.e. eating, preparing, acquiring). Each meta-class is in turn split into sub-classes, until a three-level taxonomy is reached. A deep Convolutional Neural Network (CNN) is applied to each level of the taxonomy, resembling a DECOC classifier \cite{decoc}, which decomposes a multi-class classification problem into multiple classification problems organized hierarchically. The model proposes not only the classification into 15 food-related scene classes corresponding to the lowest level of the taxonomy, but also the recognition of meta-classes at different levels (e.g. cooking, shopping, eating), which provides a more general view on the nutritional behaviour. However, the classification of images does not give a higher level of understanding about the behaviour of the person. In this work, we go a step further by giving tools for the automatic analysis of the given labels for nutritional habits discovery. 

Moreover, previous works on routine discovery found that anomaly detection methods were able to identify days considered as non-routine related \cite{caip}. We build on top of these findings for the discovery of nutritional habits and routines from unseen photo-streams. 

From another perspective, food detection and recognition from images have been widely studied. For example, in \cite{kagaya2014food}, the authors focused on the recognition and tracking of food-intake in images where food occupies a significant part of the image. These approaches are not suitable for our research since they just focus on classifying single images. In our case, we are dealing with continuous sets of labels that describe the lifestyle of the user.

To the best of our knowledge, this is the first work that addresses the analysis and discovery of nutritional habits from egocentric photo-streams. 

\section{Methodology} 
\label{sec:methodology}

Our goal is to identify the food-related habits of a person by studying the food-related environment where he or she spends time throughout the day. Firstly, we describe our proposed pipeline for the classification of unseen photo-streams, which improves the state-of-the-art \cite{hierarchical}. Secondly, we depict the model for nutritional habits discovery, which relies on the assigned labels to food-related images.

\begin{figure}[h!]
    \centering
    \includegraphics[width=1.0\columnwidth]{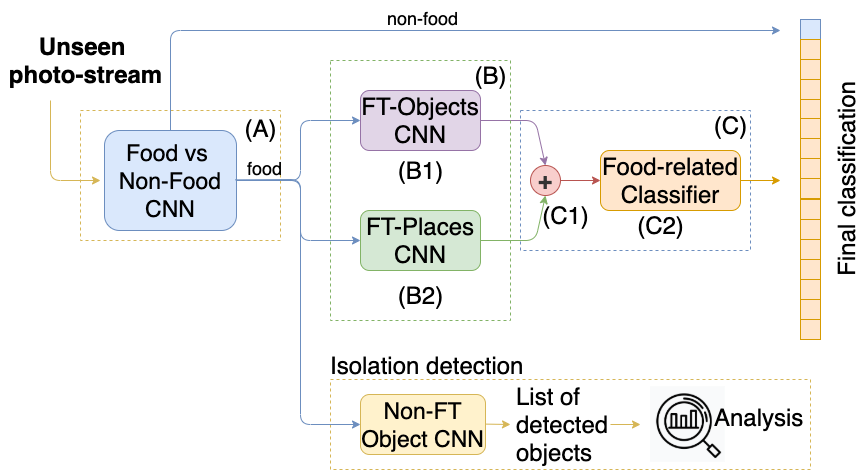}
    \caption{Proposed image classification pipeline. The model accepts as input an unseen photo-stream, determines which are the food and non-food scenes and provides an appropriate classification for the food-related scenes found. The model is extended for the identification of isolated scenes by analyzing detected objects in food-related images.}
    \label{fig:pipeline}
\end{figure}

\subsection{Food-related Scenes Classification}

For the recognition of food-related images we propose a pipeline inspired by the approach proposed in \cite{objscene}. The authors have shown that the classification accuracy was improved when combining the output of two networks, one trained for the classification of places and the other for objects. In our approach, we rely on pre-trained networks on places and objects. We further train these networks on the food scenes that compose the \textit{EgoFoodPlaces} dataset \cite{hierarchical}. 

The addition of a binary classification block for food vs non-food related image classification is essential for the automatic analysis of unseen photo-streams. It goes without saying that there is need of a robust set of non-food related images to balance the wide range of food-related samples in the \textit{EgoFoodPlaces} dataset. Even though errors at this stage of the classification pipeline can easily propagate, its addition is crucial for a real applicability of the model. An overview of the architecture of the pipeline is shown in Fig. \ref{fig:pipeline}.


\begin{enumerate}[label=\Alph*]

\item \textit{Food vs. Non-food binary classification}

The \textit{food} and \textit{non-food} related images classification pipeline takes as input a set of unseen egocentric images taken chronologically throughout one day. For each image, 
of the classification, whether it is a food scene or not is determined using the binary classifier. In the case the input image is classified as food scene, it is further passed to the next stages in the classification pipeline to be labeled according to the 15 food scene classes in the dataset. In the case the input is not considered a food scene, the label non-food is attributed and the classification process ends. 

\item \textit{Images' description through detected objects and context}

In \cite{GonzalezDiaz2013Modeling}, the authors argued that activities involve sequences of active objects and places where they are performed. In this work, we explore this idea and describe images based on semantics that describe objects that are close to the camera wearer and the environment.

During this stage of the classification process, the input images is related to a food scenes. We rely on fine-tuned networks previously trained on object (component (B1)) and on places (component (B2)), for the characterization of images. We use the probability vector of the \textit{Softmax} layers as image descriptor. This layer is a high level representation of the semantic relation among the classes that describe the image. This is done in parallel. 

\item \textit{Food-related image classification}

During the final stage of the classification pipeline,
the results of the object and places CNNs (i.e. components (B1) and (B2), respectively) are combined and the resulting class fusion vector is passed to a traditional classifier (see component (C2)), in order to achieve the final classification. The class score fusion (see component (C1)) is computed using equation (\ref{eq1}):
  \begin{equation}
      s(I) = \alpha_o s_o (I) + \alpha_p s_p (I),
      \label{eq1}
  \end{equation}
where $s_o$ and $s_p$ are the resulting probability vectors of the object and places CNNs on image I, respectively. $\alpha_o$ and $\alpha_p$ represent the weight of both results. In our experiments, we give the same relevance to bot values, and thus $\alpha_o=\alpha_p=1$.

\end{enumerate}

As a result, an egocentric image is described with a normalized vector of features described by a 16 probability values, which represents its likelihood of representing one of the 15 food-related classes and the non-food related class.

\subsection{Nutritional Routine Discovery}

\begin{figure}[h!]
    \centering
    \includegraphics[width=1.0\columnwidth]{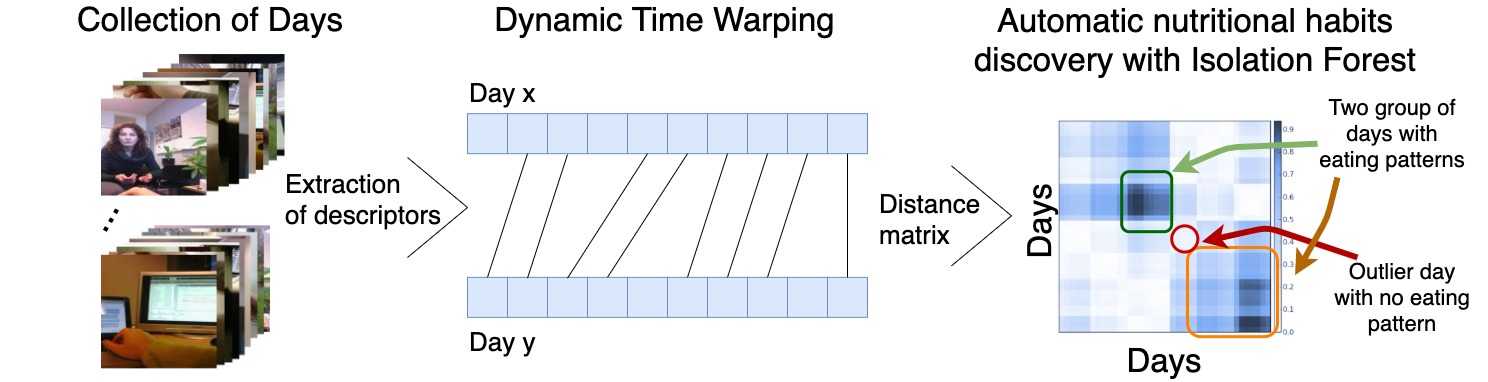}
    \caption{Proposed pipeline for the discovery of days with similar nutritional habits given a collection of recorded photo-streams.}
    \label{fig:pipelineNutritional}
\end{figure}

Given the previous classification of the collected photo-streams by a user, we analyse his or her eating habits. For the proposed \textit{Nutritional habit discovery}, we propose the pipeline presented in Fig. \ref{fig:pipelineNutritional}. First, we represent the nutritional activities within a day as a set of time-slots. A time-slot is described as the aggregation of the descriptors of the images that compose it. Therefore, a day is described by the concatenation of the descriptors for each time-slot. In our model, the time slots used are of 1 hour each, considering that 1h provides a faithful representation of the day analysed. This avoids the non-food class dominating or miss-classifications overtaking the classification, in the case of longer or shorter time slots, respectively.

Later, Dynamic Time Warping (DTW)\cite{dtw} is applied to compare the time-slots vectors that represent the different days of the user and find the correspondence of the photostreams of the days (note that people not always used to have lunch at the same time).  DTW is an algorithm for computing the optimal alignment of time series by shrinking or stretching one of the series non-linearly along the time axis such that it fits the other series used in the comparison. DTW allows for small temporal differences between the time series, which is desired for the problem at hand since daily events take place recurrently within a time margin. Given two vectors, $s'$ and $s''$, each representing a day, we can compute the path warping $w = (w_0, w_1, ..., w_Q)$, where $Q$ is the length of the path. Each element in the path $w_q$ is a tuple $(w_q[1], w_q[2])$ which indicates the mapping between the two sequences (i.e. element $w_q[1]$ in $s'$ corresponds to element $w_q[2]$ in $s''$). Equation (\ref{eq3}) describes the formula behind the DTW algorithm which computes the optimal path warping (i.e. minimal distance path) which defines the best correspondence between the sequences:
\begin{equation}
    dist_{DTW}(s', s'') = \sum_{q = 0}^{Q}dist(s'_{w_q[1]}, s''_{w_q[2]}).
    \label{eq3}
\end{equation}
 
Fast DTW \cite{fastdtw} is a linear approximation of the original DTW algorithm. For computing the distances between the time slots without loss of generality, we apply Euclidean distance. Given two nutritional timelines describing two days, the Fast DTW algorithm outputs the cost of aligning the two time series (i.e. their similarity). This method is applied for all possible pair combinations of days recorded for a user such that the overall nutritional routine can be identified.

In order to correctly group days that exhibit the same nutritional habits, we rely on the unsupervised outlier detection algorithm Isolation Forest \cite{iso_forest}. This method is applied on the results achieved by the Fast DTW. According to \cite{iso_forest_paper}, the Isolation Forest algorithm has shown promising results in the field of egocentric image data analysis. The algorithm is a tree ensemble method that randomly chooses a feature for which it selects a random split value between the possible minimum and maximum value. By recursive random partitions, the algorithm creates a tree like structure; since outlier data is assumed to be more sparse than regular data, the paths that are shorter (i.e. closer to the root) are more likely to identify the anomalies. Furthermore, an anomaly score is computed based on the average and normalized distance of the path; equation (\ref{eq4}) computes the anomaly score $a(x, n)$ for observation $x$, given a set of $n$ samples:

\begin{equation}
    a(x, n) = 2^{\frac{-E(h(x))}{c(n)}},
    \label{eq4}
\end{equation}
where $h(x)$ is the path length of point $x$ from the root node to the last external node; $E(h(x))$ identifies the average of $h(x)$ from a collection of isolation trees. The average path length is denoted by $c(n)$ following equation (\ref{eq5}): 

\begin{equation}
\begin{array}{c}
      c(n) = 2H(n-1) - \frac{2(n-1)}{n}, \\
      H(n) = ln(n) + \gamma \ (Euler's \ constant)
\end{array}
\label{eq5}
\end{equation}

We would like to highlight that the Isolation Forest algorithm allows the identification of multiple clusters. This is of importance since nutritional habits could be defined by multiple behavioural patterns. Therefore, this accounts for the fact that a person might have multiple types of routine behaviour. In this manner, the nutritional habits detection is not limited to a pre-defined number of possible groups (as opposed to other unsupervised methods that require as prerequisite the number of clusters). We believe it is relevant that we are able to capture slight changes in groups that belong to nutritional routines.

To summarize the proposed model, first, given a collection of photo-streams as input, each image is classified into food and non-food related scenes. Second, photo-streams are described as a sequence of labels, used for the later computation of similarity among days. Finally, the day results are clustered based on similarity among nutritional descriptors.

\textit{Application - Isolation identification in food scenes:} We propose to extend the proposed food and non-food classification model to keep track of isolated eating scenes (i.e. eating in front of the television or a computer). To this end, we add to the proposed pipeline a preliminary non fine-tuned CNN pre-trained on ImageNet\cite{imagenet} as seen in Fig. \ref{fig:pipeline}. Using the labels provided by ImageNet, this CNN can be used to count the number of occurrences of objects such as `laptop', `television', or `computer screen' in images that have been labeled as food scenes by the binary classifier. By identifying this type of objects in an image already determined to be a food scene by the binary classifier, we can safely assume a scenario of isolated eating occurred. For this task, we are relying on the performance of the ImageNet fine-tuned CNN, since our dataset does not include any label information about eating in isolation. 


\section{Experiments}
This section discusses the dataset used, as well as the experimental setup and the validation metrics applied to compare the achieved results.

\subsection{Dataset} \label{datset}

In this work, we make use of the \textit{EgoFoodPlaces} \cite{hierarchical} and \textit{EgoRoutine} \cite{talavera2020topic} for the training of the food vs non-food related classifier and the evaluation of the nutritional habits discovery, respectively.

On one side, for the training  of our model, we extend the existent \textit{EgoFoodPlaces} dataset with a set of 34084 new images describing non-food related scenes. Our aim is to automatically analyse sets of unseen photo-streams. Therefore, we extend the \textit{EgoFoodPlaces} dataset because it is just composed of food-related scenes, not being an accurate representation of what happens in the daily life of people. Our new collected images were manually selected and were also recorded with the Narrative Clip wearable camera. \footnote{The extended dataset will be made publicly available together with this paper.} 

The \textit{EgoFoodPlaces} dataset contains over 33000 labeled egocentric images of food-related activities of 11 users, captured with a wearable camera from a $1^{st}$ person perspective. It contains 15 low level classes and 3 meta-classes as proposed by the taxonomy in \cite{hierarchical}. An overview of the taxonomy is shown below:
\begin{itemize}
    \item Eating: Classes describing scenes where eating and/or drinking occurs: picnic area, bar, beer hall, pub indoor, cafeteria, coffee shop, dining room, and restaurant.
    \item Preparing: This class refers to cooking and it is composed by images describing the kitchen scene.
    \item Acquiring: Classes referring to scenes where food or food-related items are collected: supermarket, market outdoor, bakery, food court, market indoor, ice cream parlour.
\end{itemize}

For the training of the binary classifier, our extended \textit{EgoFoodPlaces} dataset was split in a 2:1:1 ratio between the train, validation and test sets. The food and non-food macro-classes are balanced. However the food class consists of 15 sub-classes from the \textit{EgoFoodPlaces} dataset which are highly imbalanced as shown in Table \ref{tab:class_distribution}. To mitigate this problem for the training set, a sampler was used which over sampled the classes with a lower sample count and, in turn, under sampled the classes with a higher sample count. 

\begin{table}[ht!]
\Large
\resizebox{1.0\columnwidth}{!}{
\begin{tabular}{l||c|c|c|c|c|c|c|c|c|c|c|c|c|c|c||c}
\multicolumn{17}{c}{EgoFoodPlaces Dataset}\\
\multicolumn{17}{c}{}\\
Class     & \rotatebox[origin=c]{90}{Bakery shop} &\rotatebox[origin=c]{90}{Bar} & \rotatebox[origin=c]{90}{Beer Hall} & \rotatebox[origin=c]{90}{Pub indoor} & \rotatebox[origin=c]{90}{Cafeteria} & \rotatebox[origin=c]{90}{Coffee Shop} & \rotatebox[origin=c]{90}{Dining room} & \rotatebox[origin=c]{90}{Food court} & \rotatebox[origin=c]{90}{IceCream Parlour} & \rotatebox[origin=c]{90}{Kitchen} & \rotatebox[origin=c]{90}{Market indoor} & \rotatebox[origin=c]{90}{Market outdoor} & \rotatebox[origin=c]{90}{Picnic area} & \rotatebox[origin=c]{90}{Restaurant} & \rotatebox[origin=c]{90}{Supermarket} &\rotatebox[origin=c]{90}{ \#Food Images} \\
\hline
\hline

Train       & 103         & 1077 & 296       & 342        & 1122      & 1538        & 2465        & 180        & 70               & 2635    & 902           & 1115           & 659         & 7168       & 3659       &  23331 \\
\hline
Test     & 26          & 368  & 62        & 109        & 405       & 539         & 738         & 6          & 25               & 837     & 188           & 163            & 173         & 2131       & 853        & 6623 \\
\hline
Val  & 15          & 187  & 314       & 60         & 162       & 236         & 436         & 18         & 12               & 365     & 91            & 110            & 89          & 1011       &    750   & 3856 \\
\hline 
\begin{tabular}[c]{@{}l@{}}Total\end{tabular} & 144 & 1632 & 672 & 511 & 1689 & 2313 & 3639 & 204 & 107 & 3837 & 1181 & 1388 & 921 & 10310 & 5262 & \textbf{33810}\\ \hline 
\multicolumn{17}{c}{}\\
\multicolumn{17}{c}{Binary Food \& Non-food Dataset}\\

Class     & \multicolumn{6}{c|}{Food} & \multicolumn{6}{c|}{Non-food} & \multicolumn{4}{c}{Total per set}\\
\hline \hline
Train     & \multicolumn{6}{c|}{23236} & \multicolumn{6}{c|}{23537} & \multicolumn{4}{c}{46773 }\\
\hline
Test     & \multicolumn{6}{c|}{1027} & \multicolumn{6}{c|}{7460} & \multicolumn{4}{c}{8487}\\
\hline
Val     & \multicolumn{6}{c|}{6596} & \multicolumn{6}{c|}{6280} & \multicolumn{4}{c}{12876}\\
\hline
Total     & \multicolumn{6}{c|}{30859} & \multicolumn{6}{c|}{37277} & \multicolumn{4}{c}{68136}\\
\hline

\end{tabular}

}
\caption{Distribution of images per non-food and food scene class and per train, test and validation sets.}
\label{tab:class_distribution}
\end{table}

For the purpose of testing the proposed model for nutritional habits discovery we use the unlabeled \textit{EgoRoutine} dataset. \textit{EgoRoutine} is a collection of egocentric images captured by 7 different users over approximately 15 days which illustrates the users' daily activities. An overview of the dataset distribution is shown in Table \ref{tab:routine_users}.

\begin{table}[h!]
    \centering
    \resizebox{\columnwidth}{!}{
    \begin{tabular}{c|c|c|c|c|c|c|c}
         & User 1& User 2 & User 3 & User 4 & User 5& User 6& User 7 \\
         \hline \hline
         \#Days & 14 & 10 & 16 & 21 & 13 & 18 & 13\\
         \#Images & 20543 & 11815 & 21727 & 18977 & 17046 & 16592 & 11207\\
         \hline
    \end{tabular}
    }
    \caption{Distribution of \textit{EgoRoutine} dataset presenting the number of days recorded and the total number of images recorded by each user in the dataset.}
    \label{tab:routine_users}
\end{table}

\begin{figure}[h!] 
    \centering
    \includegraphics[width=\columnwidth]{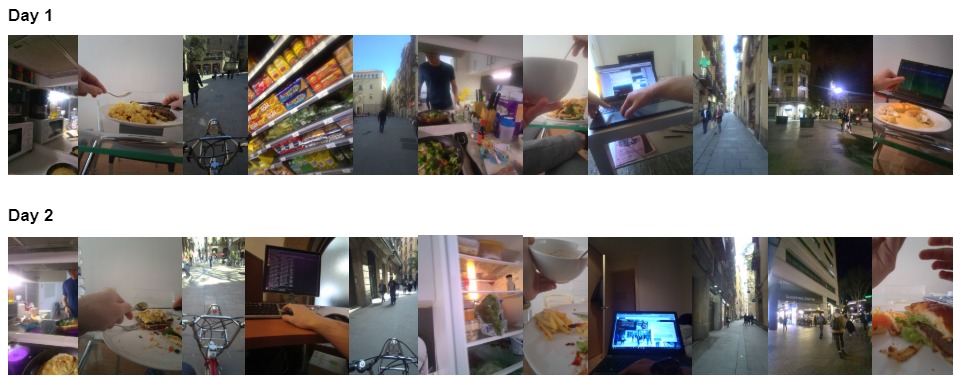}
    \caption{Examples of images from two recorded days in the \textit{EgoRoutine} dataset.}
    \label{fig:photo_stream}
\end{figure}

As expected, these images mostly depict non-food environments, but also contain some food scenes amongst them, as shown in Fig~\ref{fig:photo_stream}.

\subsection{Validation}

We use different metrics for the evaluation of the performance of our model. We rely on \textit{Accuracy} (Acc), \textit{Precision} (P), \textit{Recall} (R), and \textit{F-score} to assess the performance of the proposed food and non-food related classifier. 
Moreover, we report the macro (M) and the weighted (W) accuracy. The W accuracy gives a more accurate insight since it considers the imbalanced nature of the classes. The validation of the clustering is measured by applying the Silhouette score metric \cite{silhouette}. This metric describes the cohesion for each point w.r.t. its assigned cluster, and is defined by Equation (\ref{eq:silh}), where $a(i)$ is the average distance between point $i$ and points withing the same cluster, and $b(i)$ is the minimum average distance from $i$ to points in the other clusters:
\begin{equation}
Silhouette_{score} = \frac{b(i) - a(i)}{max(a(i), b(i))}\label{eq:silh}.
\end{equation}

\subsection{Experimental Setup}
In this sub-section, we present the experimental setup for the proposed methods for both the \textit{food-related scene classification} and the \textit{nutritional habits discovery}.

\subsubsection{Food-related scenes classification}

We perform an ablation study where the classifiers described in Section \ref{sec:methodology} are implemented as CNNs. With these auxiliary methods, we are aiming to create a solid basis of comparison for the proposed method. The experiments conducted for the food and non-food related images classification were defined as follows for the different components of the pipeline presented in Fig~\ref{fig:pipeline}:

\begin{enumerate}[label=\Alph*]

\item \textit{Food vs Non-food binary classification:} 
We evaluate the performance of the ResNet50 \cite{resnet}, DenseNet161 \cite{densenet}, and VGG16 \cite{vgg} for the task of binary classificaiton of images into food and non-food scene related. These architectures were pre-trained on ImageNet \cite{imagenet} and further fine-tuned on our extended version of the \textit{EgoFoodPlaces} dataset, which is composed of food and non-food images. 

\item \textit{Food-related images description through the detection of objects and context:} 

\begin{enumerate}

\item Object detector: ResNet50 \cite{resnet}, DenseNet161 \cite{densenet}, VGG16\cite{vgg}, GoogLeNet Inception V3 architectures pretrained on ImageNet \cite{imagenet}.

\item Place recognition: ResNet50 \cite{resnet} pre-trained on Places365 \cite{places365}.

\end{enumerate}

For all the architectures used (including the binary classifier), fine-tuning was required to adapt to the problem at hand. Thus, each model was further trained on the \textit{EgoFoodPlaces} dataset for 10 epochs, with a 0.001 learning rate.

\item \textit{Food-related images classification:}
\begin{enumerate}

\item Class score fusion: we evaluate the performance of the model by aggregating or concatenating the probability vectors resulted from the object detector and place recognition CNNs.
\item Traditional classifiers:  Support Vector Machine (SVM) \cite{svm}, Random Forest (RF)(200 trees)\cite{rf} and K-nearest neighbours (KNN)(8 neighbours) \cite{knn}.
\end{enumerate}

\end{enumerate}

As described above, we mainly experimented with different already pre-trained architectures which we fine-tune over the \textit{EgoFoodPlaces} dataset. 

\subsubsection{Nutritional routine discovery}
Routines of different persons are different; hence we cannot use supervised learning techniques to classify nutritional routine vs no routine. In order to discover the routine as cluster of day with similar nutritional behaviour, we test two  clustering approaches to group the results of the DTW algorithm with the aim to identify which days follow a nutritional routine. For this purpose, we employ two distinct clustering methods: Hierarchical Agglomerative Single-linkage clustering\cite{singlelink} and Density Based Spatial clustering (DBscan)\cite{dbscan}. Single-linkage clustering gradually combines clusters based on the minimum distance between pairs of points not belonging to the same cluster, while DBscan groups together points that belong to dense regions, marking as outliers points that belong to low density regions.


\section{Results} 
\label{Section4:results} 
In this section, we present the obtained results of our proposed model. On one side, we discuss the introduced pipeline for the food and non-food related image classification. On the other side, we examine the discovered nutritional patterns and what we can deduce from them.

\subsection{Results for the \textit{food} and \textit{non-food} related images classification}

\textit{Image analysis:} We present the classification performance at image label of the baseline models without a pre-pended binary classifier in Table \ref{table:overview_var1}. We observe that models using the KNN traditional classifier and aggregation of the classification results generally achieve an overall higher score. We can highlight the performance of the following architectures: the DenseNet + Places (aggregated) with KNN and the GoogLeNet + Places (aggregated) with KNN, which obtain a W accuracy of 70\% and 69\% and a F-score of 63\% and 64\%, respectively


\begin{table}
\resizebox{0.5\textwidth}{!}{
\begin{tabular}{l|l|l|l|l|l|l|l|l|l}
\small
\begin{tabular}[c]{@{}l@{}}Traditional\\ classifier\end{tabular} 
  & \begin{tabular}[c]{@{}l@{}}Fine-tuned\\ architectures\end{tabular}   & \multicolumn{2}{c|}{ Acc} & \multicolumn{2}{c|}{ F-score } & \multicolumn{2}{c|}{ P } & \multicolumn{2}{c}{ R } \\
& & M & W & M & W & M & W & M & W \\
\hline
\hline
\multirow{8}{*}{ SVM } & ResNet + Pl (concat) & 0.52 & 0.51 & 0.18 & 0.43 & 0.40 & 0.63 & 0.18 & 0.52 \\
& ResNet + Pl (aggreg) & 0.33 & 0.32 & 0.03 & 0.16 & 0.20 & 0.43 & 0.06 & 0.32 \\
& DenseNet + Pl (concat) & 0.50 & 0.52 & 0.17 & 0.41 & 0.41 & 0.63 & 0.16 & 0.49 \\
& DenseNet + Pl (aggreg) & 0.32 & 0.33 & 0.03 & 0.16 & 0.18 & 0.39 & 0.06 & 0.32 \\
& VGG + Pl (concat) & 0.48 & 0.47 & 0.15 & 0.39 & 0.41 & 0.63 & 0.15 & 0.47 \\
& VGG + Pl (aggreg) & 0.33 & 0.31 & 0.03 & 0.16 & 0.26 & 0.51 & 0.06 & 0.32 \\
& GoogLeNet + Pl (concat) & 0.64 & 0.65 & 0.33 & 0.58 & 0.67 & \textbf{0.70} & 0.30 & 0.64 \\
& GoogLeNet + Pl (aggreg) & 0.51 & 0.52 & 0.18 & 0.43 & 0.53 & \textbf{0.71} & 0.17 & 0.51 \\
\multirow{8}{*}{ KNN } & ResNet + Pl (concat) & 0.66 & 0.66 & 0.44 & 0.61 & 0.62 & 0.64 & 0.40 & 0.64 \\
\hline
& ResNet + Pl (aggreg) & 0.67 & 0.67 & 0.45 & 0.63 & 0.65 & 0.66 & 0.41 & 0.67 \\
& DenseNet + Pl (concat) & 0.66 & \textbf{0.70} & 0.44 & 0.62 & 0.65 & 0.66 & 0.40 & 0.66 \\
& \textbf{DenseNet + Pl (aggreg)} & \textbf{0.68} & \textbf{0.70} & 0.46 & 0.63 & 0.64 & 0.67 & 0.42 & 0.67 \\
& VGG + Pl (concat) & 0.66 & 0.65 & 0.44 & 0.61 & 0.63 & 0.64 & 0.40 & 0.66 \\
& VGG + Pl (aggreg) & 0.66 & 0.64 & 0.44 & 0.62 & 0.62 & 0.65 & 0.40 & 0.65 \\
& GoogLeNet + Pl (concat) & \textbf{0.68} & 0.68 & 0.46 & 0.63 & 0.65 & 0.67 & 0.42 & 0.67 \\
& GoogLeNet + Pl (aggreg) & \textbf{0.68} & 0.69 & \textbf{0.48} & \textbf{0.64} & \textbf{0.67} & 0.68 & \textbf{0.43} & \textbf{0.68} \\
\multirow{8}{*}{ RF } & ResNet + Pl (concat) & 0.65 & 0.64 & 0.42 & 0.61 & 0.57 & 0.63 & 0.39 & 0.64 \\
\hline
& ResNet + Pl (aggreg) & 0.66 & 0.67 & 0.42 & 0.62 & 0.55 & 0.64 & 0.39 & 0.66 \\
& DenseNet + Pl (concat) & 0.65 & 0.68 & 0.44 & 0.61 & 0.58 & 0.63 & 0.41 & 0.65 \\
& DenseNet + Pl (aggreg) & 0.66 & \textbf{0.70} & 0.44 & 0.62 & 0.58 & 0.64 & 0.41 & 0.66 \\
& VGG + Pl (concat) & 0.63 & 0.61 & 0.42 & 0.59 & 0.55 & 0.61 & 0.38 & 0.62 \\
& VGG + Pl (aggreg) & 0.64 & 0.63 & 0.42 & 0.60 & 0.59 & 0.63 & 0.38 & 0.64 \\
& GoogLeNet + Pl (concat) & 0.65 & 0.65 & 0.43 & 0.61 & 0.57 & 0.63 & 0.40 & 0.64 \\
& GoogLeNet + Pl (aggreg) & 0.67 & 0.67 & 0.44 & 0.63 & 0.56 & 0.66 & 0.41 & 0.67 \\
\hline
\end{tabular}
}

\caption{Results of the performed ablation study for the classification of images into food-related scenes. 'Aggreg' stands for aggregation and 'concat' for concatenation. 
}
\label{table:overview_var1}
\end{table}
Table \ref{tab:res_binary} presents the results achieved by the different architectures for the binary classifier of the \textit{food} and \textit{non-food} related images classification pipeline. As shown in Table \ref{tab:res_binary}, VGG16 achieves the highest score for all the metrics.
\begin{table}

\resizebox{\columnwidth}{!}{\begin{tabular}{l|l|l|l|l|l|l|l|l}
    Binary classifier & \multicolumn{2}{c}{ Accuracy } & \multicolumn{2}{c}{ F-score } & \multicolumn{2}{c}{ Precision } & \multicolumn{2}{c}{ Recall } \\
    architecture & macro & weighted & macro & weighted & macro & weighted & macro & weighted \\
    \hline
    \hline
    ResNet & 0.74 & 0.74 & 0.72 & 0.72 & 0.77 & 0.77 & 0.73 & 0.74 \\
    DenseNet & 0.77 & 0.77 & 0.77 & 0.76 & \textbf{0.81} & \textbf{0.81} & 0.78 & 0.77 \\
    \textbf{VGG} & \textbf{0.80} & \textbf{0.80} & \textbf{0.80} & \textbf{0.80} & \textbf{0.81} & \textbf{0.81} & \textbf{0.80} & \textbf{0.80} \\
    \hline
    \end{tabular}}
    \caption{Performance of the food vs non-food related image classifiers. 
    }
    \label{tab:res_binary}

\end{table}

\textit{Photo-streams analysis:}
Table \ref{table:res_tool1} presents the results of the different experiments conducted for the food and non-food related images classification pipeline. We compare our pipeline method using DenseNet architecture to the GoogLeNet architecture which had a comparable performance at the image label level (see Fig~\ref{table:overview_var1}).
As it can be observed, the proposed pipeline outperforms the baseline methods, by obtaining an overall accuracy and $W$ F-score of 61\%. Classes with the highest accuracy are 14, supermarket, 13, restaurant, 10, market indoor, with accuracies of $84\%$, $73\%$ and $73\%$, respectively, while classes 7, food court, and 12, picnic area, fail to be classified. This can be explained by the limited amount of samples from these two classes as shown in Table \ref{tab:class_distribution}.


\begin{table}
\resizebox{0.5\textwidth}{!}{
\begin{tabular}{l|cc|cc}
& \multicolumn{2}{c}{ KNN } & \multicolumn{2}{c}{ RF } \\
& \textbf{Our Method} & GoogleNet (A) & DenseNet (A) & GoogleNet (A)\\
\hline \hline
0 (bakery shop) & 0.43 & 0.53 & 0.36 & 0.37 \\
1 (bar) & 0.12 & 0.25 & 0.09 & 0.32 \\
2 (beer hall) & 0.25 & 0.22 & 0.23 & 0.26 \\
3 (pub indoor) & 0.57 & 0.54 & 0.48 & 0.32 \\
4 (cafeteria & 0.48 & 0.49 & 0.48 & 0.47 \\
5 (coffee shop) & 0.42 & 0.49 & 0.40 & 0.39 \\
6 (dining room) & 0.62 & 0.57 & 0.64 & 0.57 \\
7 (food court) & 0.00 & 0.00 & 0.00 & 0.00 \\
8 (ice cream parlor) & 0.53 & 0.68 & 0.44 & 0.39 \\
9 (kitchen) & 0.70 & 0.70 & 0.70 & 0.69 \\
10 (market indoor) & 0.73 & 0.68 & 0.76 & 0.56 \\
11 (market outdoor) & 0.17 & 0.24 & 0.2 & 0.22 \\
12 (picnic area) & 0.00 & 0.00 & 0.00 & 0.00 \\
13 (restaurant) & 0.73 & 0.70 & 0.70 & 0.63 \\
14 (supermarket) & 0.84 & 0.8 & 0.85 & 0.79 \\
\hline
Accuracy & \textbf{0.61} & 0.59 & \textbf{0.60} & 0.55 \\
Precision (macro) & \textbf{0.62} & 0.54 & 0.59 & 0.52 \\
Precision (weighted) & \textbf{0.69} & 0.65 & 0.67 & 0.62 \\
Recall (macro) & 0.35 & \textbf{0.39} & 0.34 & 0.33 \\
Recall (weighted) & \textbf{0.61} & 0.59 & \textbf{0.60} & 0.55 \\
F-score (macro) & \textbf{0.41} & 0.43 & \textbf{0.40} & 0.37 \\
F-score (weighted) & \textbf{0.61} & \textbf{0.61} & \textbf{0.60} & 0.57 \\
\hline
\end{tabular}}

\caption{Performance of the pipeline for \textit{food} and \textit{non-food} related images classification for unseen egocentric photo-stream. 
}

\label{table:res_tool1}
\end{table}

\begin{figure}
\centering
\includegraphics[width=\columnwidth]{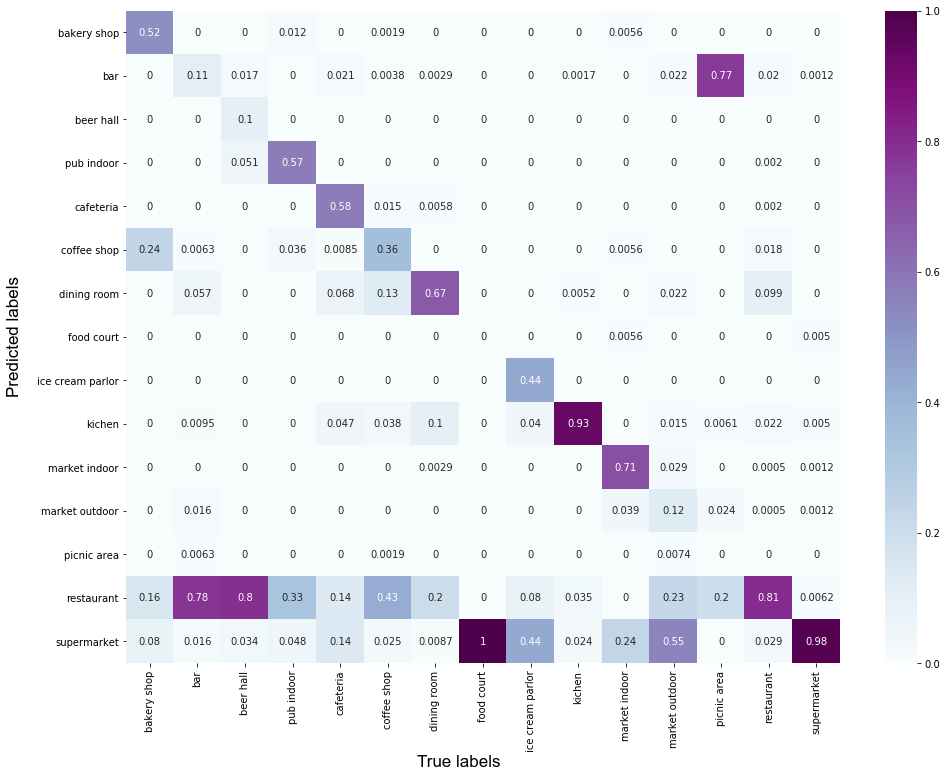}
\caption{Confusion matrix showing the classification performance of the food related images. 
}
\label{fig:confusionmatrix}
\end{figure}

The $W$ metric shows the better performance of our proposed food and non-food related images classification method w.r.t. the more complex hierarchical classifier model described in \cite{hierarchical}, which achieved a final $W$ accuracy and F-score of 56\% and 65\%, respectively. In Fig~\ref{fig:confusionmatrix}, the confusion matrix shows the distribution of miss-classifications. We can observe how classes with less amount of samples tend to be miss-classified and that class 'restaurant' tends to be mainly incorrectly classified to classes that show social eating (i.e. `bar', `beer hall', `pub indoor', `coffee shop'). This can be accounted for the low inter-class variance between the classes that depict a social context.

\subsection{Nutritional habits discovery}

\begin{figure}
\centering
\begin{subfigure}[b]{0.4\textwidth}
   \includegraphics[width=1\linewidth]{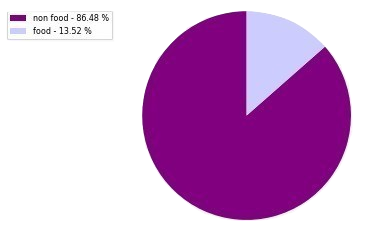}
   \caption{}
   \label{fig:User1_results1} 
\end{subfigure}
\hfill 
\begin{subfigure}[b]{0.5\textwidth}
   \includegraphics[width=1\linewidth]{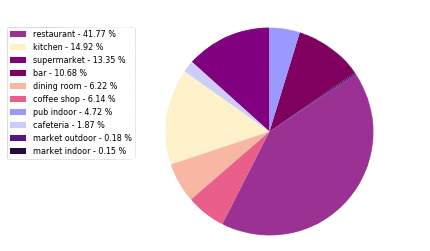}
   \caption{}
   \label{fig:User1_results1}
\end{subfigure}
\caption[Two numerical solutions]{Final classification for images collected by user 1 over all days. (a) Frequency of non-food or food-related activities. (b) Frequency of food-related activities, i.e. time spent in a food-related scene.}
\label{percentage_classes_user1}
\end{figure}

\begin{figure}[h!]
\centering
\includegraphics[width=1\columnwidth]{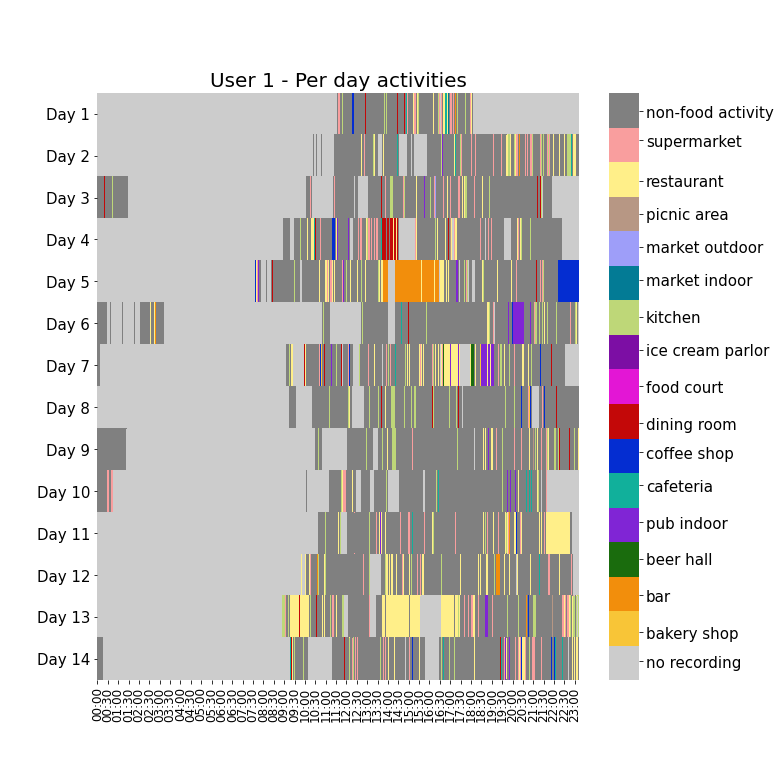}
\caption{Activity timeline over each recorded day for user 1 from the \textit{EgoRoutine} dataset. The majority class is computed per minute and displayed; in case a food scene appears during the recorded minute, the class 'non-food' is not consider in order to account for the sparsity of food scenes; this however might allow for the display of misclassifications.}
\label{fig:user1_timeline}
\end{figure}

The results of the proposed nutritional habits discovery tool are showcased through the presentation of the nutritional behaviours analysis of user 1 from the \textit{EgoRoutine} dataset. Fig. \ref{percentage_classes_user1} presents the frequency of food-related activities discovered for user 1. Only 13.52\% of the recorded time has been identified as spent in a food-related context, with class `restaurant' appearing most frequently. This can also be recognized in Fig.~\ref{fig:user1_timeline}, which shows the sparsity of the food-related scenes w.r.t. the temporal context of the recorded days. This observation can lead to the assumption that the user pays little attention to their nutritional routine. In order to corroborate this assumption, we will investigate the routine clusters discovered for user 1.

\begin{figure}[h!]
\centering
\includegraphics[width=0.8\columnwidth,height=0.7\columnwidth]{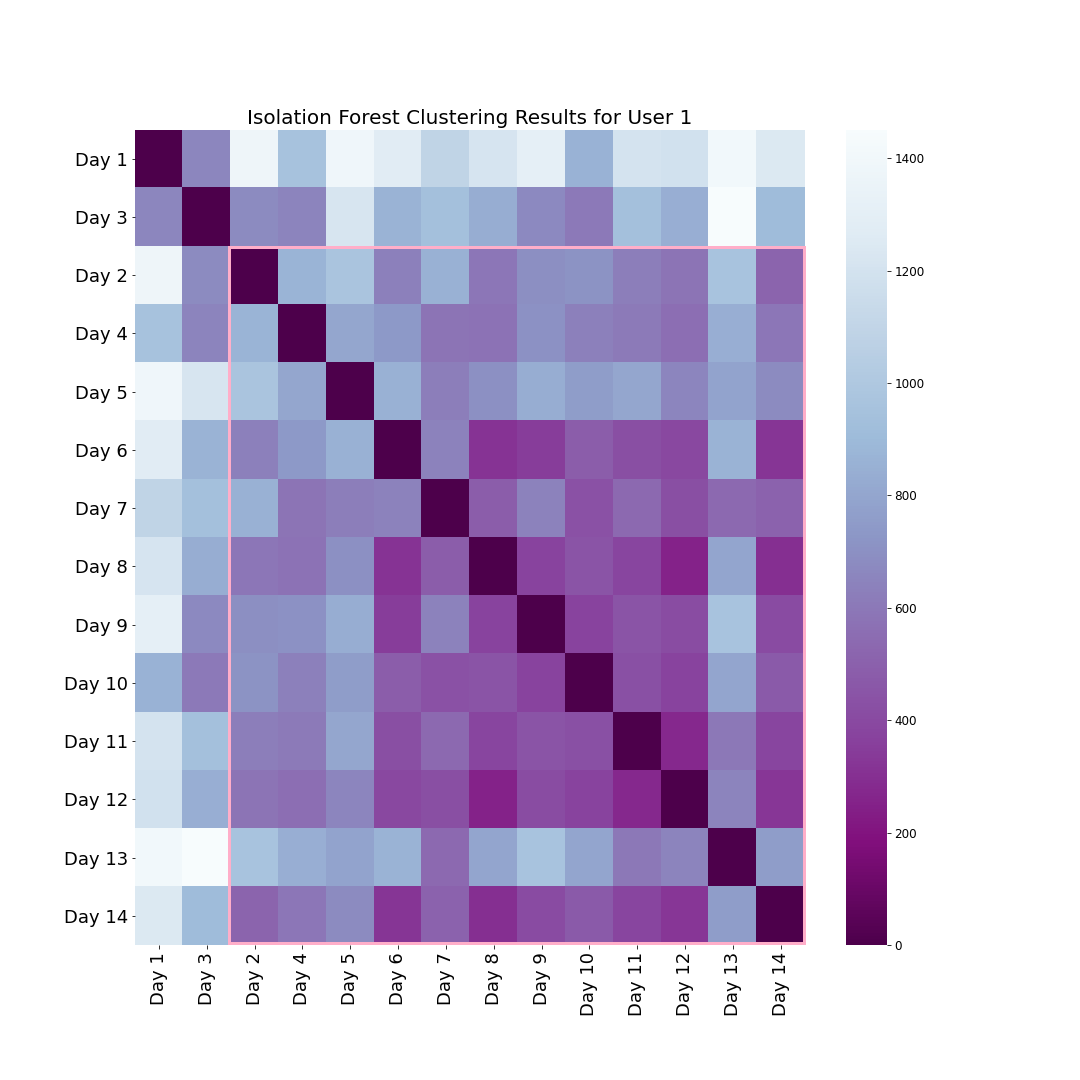}

\caption{Clustering output of Isolation Forest for the recorded data by user 1; one nutritional routine is identified, delimited by the highlighted squares. The days outside the highlighted square are considered non-routine days.}
\label{fig:user1_isoforest}
\end{figure}

An overview of the Silhouette scores obtained over the \textit{EgoRoutine} users by the three methods tested: hierarchical clustering, DBSCAN, and Isolation Forest, are showcased in Table \ref{tab:silhouette}.
\begin{table}[H]

\resizebox{1.0\columnwidth}{!}{
\begin{tabular}{l|l|l|l|l|l|l|l|l}
    Method & User 1 & User 2 & User 3 & User 4 & User 5& User 6 & User 7 & Mean \\
    \hline
    \hline
    Hierarchical & 0.251 &  0.421 & 0.134 & 0.126 & 0.178 & 0.317 & 0.203 & 0.235 \\
    DBSCAN & 0.247 & 0.478 & 0.025 & 0.172 & 0.198 & 0.317 & 0.217 & 0.223\\
    Isolation Forest & 0.477 & 0.478 & 0.272 & 0.341 & 0.156 & 0.137 & 0.222 & 0.298\\

\end{tabular}}
\caption{Silhouette score of the experiments.} 
\label{tab:silhouette}
\end{table}
We can observe that Isolation Forest produces more consistent clusters according to the overall mean score. Fig. \ref{fig:user1_isoforest} presents the results of applying the Isolation Forest method on user's 1 recorded data. Two groups were identified: the routine group, consisting of days 2, 4 and 5 through 14, and an outlier group, consisting of days 1 and 3, which we consider not to be part of the user's nutritional routine. Within the routine group, days 6 through 12 display higher similarity with each other than the remaining days in the same routine group. This could indicate the presence of sub-routines.

\begin{figure}[h!]
    \centering
    \includegraphics[width=\columnwidth]{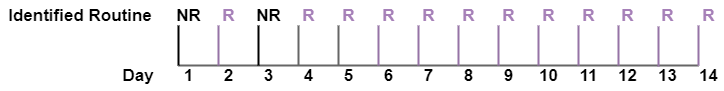}
    \caption{Discovered nutritional habits of user 1; R identify days with a certain nutritional routine, NR identifies days which follow no routine.}
    \label{fig:user1_sequencedayslables}
\end{figure}

Quantitatively, the nutritional behaviour of user 1 can be divided into two separate groups: routine and non-routine behaviour. In order to better grasp the nutritional tendencies of the user, we also need to consider the temporal aspect involved with the recorded data. Fig. \ref{fig:user1_sequencedayslables} presents the display of the routine w.r.t. all the recorded days of user 1. This overall timeline also considers the present non-routine behaviour. The timeline commences with the two identified non-routine days (i.e. days 1 and 3) disrupted by day 2 which is part of the routine group. Following day 3, user 1 adheres to a consistent routine until the last recorded day. The nutritional behaviour resented by the day sequence in Fig. \ref{fig:user1_sequencedayslables} indicates that user 1 is generally consistent with their eating habits. The two non-routine days could correspond to days in which the user did not follow their usual activities or could be regarded as 
``cheat days" during which the user disregarded their usual diet. It could also be the case that this user has a broader pattern of inconsistencies in his or her nutritional routine from which is trying to recover rapidly given the fact that the non-routine sequence spans only 2 days. However, these assumptions lacks a precise context; to confirm any of the hypotheses, recording throughout longer periods of time is needed. Unfortunately, this information is not captured by the \textit{EgoRoutine} dataset.

\begin{table*}[h!]
\Large
\resizebox{\textwidth}{!}{
\begin{tabular}{l||c|c||c|c|c|c|c|c|c|c|c|c|c|c|c|c|c||c}
\multicolumn{17}{c}{}\\
Class     & \rotatebox[origin=c]{90}{\# images}
& \rotatebox[origin=c]{90}{\# food-related}
& \rotatebox[origin=c]{90}{\% (\#) Bakery shop} &
\rotatebox[origin=c]{90}{\% (\#) Bar} & 
\rotatebox[origin=c]{90}{\% (\#) Beer Hall} & 
\rotatebox[origin=c]{90}{\% (\#) Pub indoor} & 
\rotatebox[origin=c]{90}{\% (\#) Cafeteria} & 
\rotatebox[origin=c]{90}{\% (\#) Coffee Shop} & 
\rotatebox[origin=c]{90}{\% (\#) Dining room} & 
\rotatebox[origin=c]{90}{\% (\#) Food court} & 
\rotatebox[origin=c]{90}{\% (\#) IceCream Parlour} & 
\rotatebox[origin=c]{90}{\% (\#) Kitchen} & 
\rotatebox[origin=c]{90}{\% (\#) Market indoor} & 
\rotatebox[origin=c]{90}{\% (\#) Market outdoor} & 
\rotatebox[origin=c]{90}{\% (\#) Picnic area} & 
\rotatebox[origin=c]{90}{\% (\#) Restaurant} & 
\rotatebox[origin=c]{90}{\% (\#) Supermarket} &
\rotatebox[origin=c]{90}{\% (\#) Eating isolation} \\
\hline
\hline

User1  & 20543  & 3180 & 0.35(11) & \textbf{9.03(287)} & 0.85(27) & \textbf{5.38(171)} & 2.23(71) & \textbf{5.35(170)} & 4.12(131) & 0(0) & 0(0) & \textbf{16.42(522)} & 0.09(3) & 0.19(6) & 0.28(9) & \textbf{39.53(1257)} & 16.19(515) & 8.99(286) \\
User2  & 11815  & 2576 & 0(0) & 1.01(26) & 0(0) & 2.37(61) & 0.58(15) & 0.54(14) & 6.64(171) & 0(0) & 0(0) & \textbf{32.57(839)} & 0.35(9) & 2.17(56) & 2.6(67) & \textbf{40.57(1045)} & 10.6(273) & 11.34(292) \\
User3  & 21727  & 5536 & 0.14(8) & \textbf{6.32(350)} & 0.09(5) & 3.88(215) & \textbf{10.39(575)} & 0.94(52) & \textbf{11.42(632)} & 0.18(10) & 0(0) & 11.52(638) & 0.16(9) & 0.07(4) & 0(0) & \textbf{32.39(1793)} & \textbf{22.49(1245)} & 16.24(899) \\
User4  &  18977 & 7453 & 0.03(2) & 2.05(153) & 0(0) & 1.22(91) & 0.47(35) & 0.82(61) & \textbf{10.43(777)} & 0.01(1) & 0.01(1) & \textbf{19.91(1484)} & 1.03(77) & 2.44(182) & 2.09(156)& \textbf{47.98(3576)} & 11.5(857) & 7.66(571) \\
User5  & 17046  & 5940 & 0(0) & 3.5(208) & 0(0) & 2.64(157) & 1.8(107) & 0.45(27) & 1.75(104) & 0.29(17) & 0.03(2) & 8.01(476) & 0.54(32) & 1.46(87) & 3.84(228) & \textbf{58.33(3465)} & \textbf{17.34(1030)} & 11.82(702) \\
User6  & 16592  & 3490  & 0.03(1) & 1.78(62) & 0.06(2) & 2.06(72) & \textbf{10.29(359)} & 0.63(22) & 5.19(181) & 0(0) & 0(0) & 11.78(411) & 1.49(52) & 0.49(17) & 0.72(25) & \textbf{52.95(1848)} & 12.55(438) & 9.34(326) \\
User7  & 11207  & 3532 & 0.2(7) & 3.28(116) & 0(0) & 2.72(96) & 3.51(124) & 2.6(92) & 5.38(190) & 0(0) & 0(0) & 12.34(436) & 0.31(11) & 3.62(128) & \textbf{12.74(450)} & \textbf{43.04(1520)} & 10.25(362) & 9.14(323) \\
\hline
\end{tabular}}
\caption{Given the EgoRoutine dataset, quantification of the user's nutritional-related scenes. We can observe the (\%) percentage and (\#) number of images per food-related class, of images that indicate isolation while in eating scenes. }
\label{tab:images_class_user}
\end{table*}

\begin{figure}[!h]
    \begin{minipage}[b]{0.45\textwidth}
       \begin{subfigure}[b]{0.5\linewidth}
	    \includegraphics[width=\columnwidth]{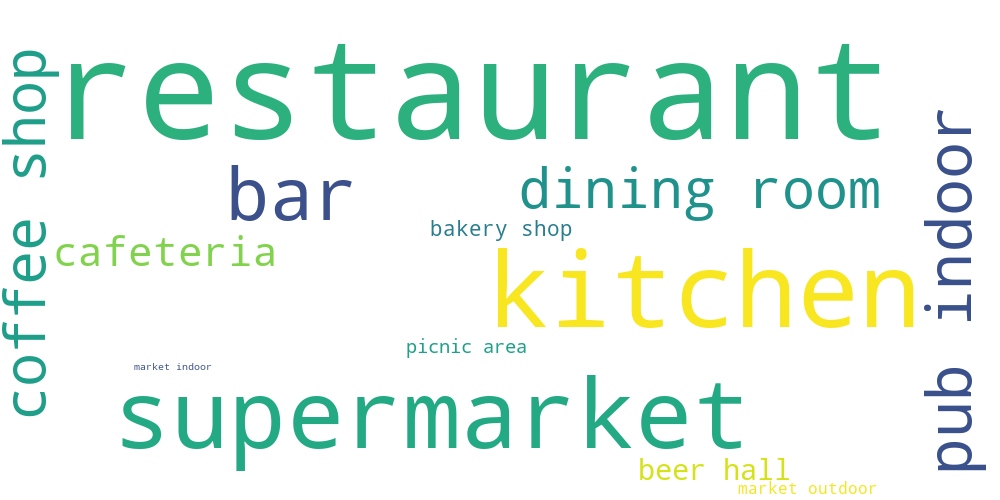}
         \caption{Routine}\label{subfig:routine2}
       \end{subfigure}
       \begin{subfigure}[b]{0.5\linewidth}
	    \includegraphics[width=\columnwidth]{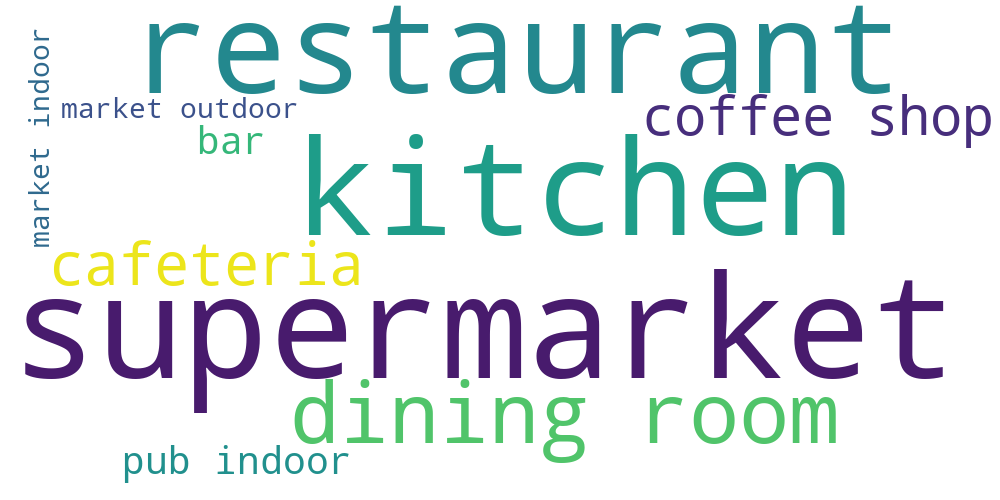}
         \caption{Non-routine}\label{subfig:nonroutine}
       \end{subfigure}
     \end{minipage}
     \hfill
     
\caption{Word cloud representation of the identified clusters by Isolation Forest for user 1. Word cloud (a) correspond to the identified routine and word cloud (b) to the remaining group of outliers which do not describe any nutritional habit. The size of the food scene names indicates the frequency of the food scene within the routine, the colours were only chosen for presentation purposes.}
\label{fig:user1_wordclouds}
\end{figure}

Analysing the issue of user 1's routine merely from a quantitative perspective does not suffice: despite a routine being followed, that routine might include unhealthy nutritional habits. We present a qualitative analysis of the discovered routine and non-routine days as word clouds (see Fig. \ref{fig:user1_wordclouds}). The word cloud representations showcase the frequency of the various food classes presented cumulatively in the respective cluster of routine or non-routine. The frequency is computed over all the days belonging to the respective routine or non-routine cluster. Class non-food was excluded from all cluster representations, as it was dominated by frequency (i.e. non-food scenes appear more frequently than food scenes) without providing any additional insight about the food-related habits.

The word clouds indicate overlap between the most common classes. Even though there is no consistent qualitative demarcation between nutritional routine and non-routine days, routine days include a more varied range of food-classes. Most of the classes with a significant frequency in the routine word cloud represent food scenes which are generally related to unhealthy eating habits: restaurant, bar, pub indoor, coffee shop. This might indicate that, despite the consistency, user 1 does not necessarily have a homogeneous or healthy eating routine. 

Even though there is no consistent qualitative demarcation between nutritional routine and non-routine days, the habits of user 1 are shifted time-wise or differ in frequency. For example, classes `restaurant' and `dining room' are common to both routine and non-routine days. However, they are more frequent to routine as seen in Fig. \ref{fig:user1_timeline}. In addition, as shown in Fig. \ref{fig:confusionmatrix}, class `restaurant' is often miss-classified to classes such as `bar', `beer hall', `pub indoor', `coffee shop', which usually imply some sort of social context. Therefore, in the case at hand, we consider class `restaurant' to not only refer to restaurant environments, but also to refer generally to social eating

\begin{figure*}[h!]
\centering
\includegraphics[width=1.6\columnwidth]{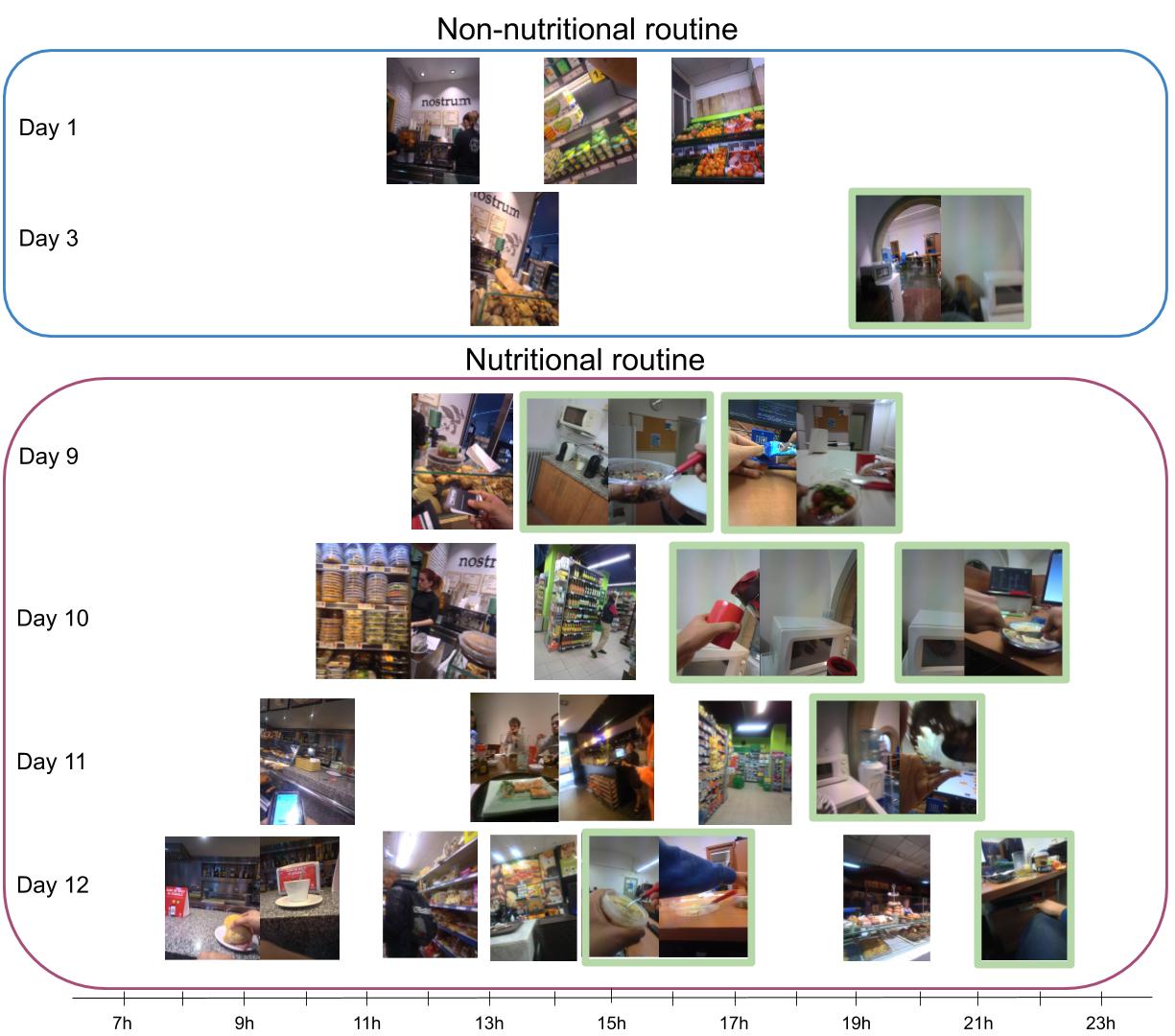}
\caption{Sample images describing recognized food-related scenes as a visual description of days that share nutritional habits for user 1. The highlighted green boxes showcase the habit of user 1 of eating and `snacking' at the office space. 
}
\label{fig:visualexmaple}
\end{figure*}

As shown in Fig. \ref{fig:user1_timeline}, the time shift is more apparent when comparing the routine and non-routine groups: non-routine days 1 and 3 are less dense in terms of the appearance of food scenes than the rest of the routine days. Moreover, the food activity is more focused at noon, while for the routine days, the user tends to spend more time in food-related environments during the evening, towards the end of the day. Another factor which differentiates the routine and non-routine days is social eating. Non-routine days 1 and 3 present fewer instances in which the user is captured in social eating environments (e.g. restaurant, bar, pub indoor), while for the routine days, social eating scenes are more common. Looking at Fig. \ref{fig:user1_timeline}, it can be observed that class `kitchen' occurs very frequently. By visually investigating the recorded images of user 1, we discovered that class `kitchen' (sometimes also 
`cafeteria' or `dining room') describes a coffee break room in the office space where the user works. Therefore, it can be inferred that the user has the habit of taking frequent breaks which include `snacking'. This explains the low percentage of only 13.52\% of the time spent involved in food activities: user 1 is in the habit of eating recurrently for very brief periods of time. In Fig. \ref{fig:visualexmaple}, we can observe sample images of days that belong to the group of days that share different nutritional habits. We can observe that similar food-related scenes appear within similar intervals of time for days that share nutritional habits. Moreover, the highlighted sequences showcase the instances of eating and `snacking' at the office space.

In Table \ref{tab:images_class_user} we present the percentage of images classified as a food-related scene per user from the \textit{EgoRoutine} dataset. These values give insight into the different behaviours when it comes to food-related habits. For example, we can observe that all users spend relatively more time eating outside in restaurants, while just user 2 spends time cooking. If focusing at a personal level, we can also deduce that user 3 eats at home often while watching tv and in isolation, since the percentages of eating in isolation and eating in the dining room are high. In contrast, user 4 spends 11\% of the time eating in isolation while almost 60\% eating at a restaurant - we can argue that s/he spends relatively a significant time eating alone outside where s/he lives, which can indicate the type of job and responsibilities of this person. If we focus on the time eating in isolation, the results indicate an average percentage of $10.65\%$ out of all the detected food scenes. The average is relatively low; considering the fact that the users spend most of their time in an office environment, in front of the computer, higher levels of isolation would be expected. However, the users seem to engage themselves in social eating, trying to avoid scenarios like eating alone in front of the computer. This hypothesis is supported by the preponderance of class restaurant, as also seen previously. With these numbers, we want to highlight that relevant conclusion can be drawn by analysing the food-related behaviour of a person.

\section{Conclusions and future work}
\label{Section6:Discussion} 

In this work, we address for the first time the automatic discovery of nutritional routine from unseen egocentric photo-streams. The proposed tools are intended to offer insights into a person's eating habits with the view of improving their daily nutritional routine towards a healthier lifestyle.

Our proposed model is capable of detecting patterns of nutritional behaviour from visual collections of data  in an unsupervised manner.  Moreover, our model allows us to quantify the discovered eating behaviours over time. It allows to draw a food intake timeline, with the corresponding class of food scene, for an entire day from a given unseen stream of egocentric images, including both food and non-food scenes.  Furthermore, we have presented an application when the model helps detecting isolation while eating, since this might be an indication of several other disorders.

Our future works on this field will study recommendation systems based on the discovered information. An end-to-end tool capable of automatically analysing the input data to later recommend actions to the user would be of high relevance for healthcare professionals and the general population.

\section*{Acknowledgment}
This work was partially founded by projects RTI2018-095232-B-C2, SGR 1742, CERCA, Nestore Horizon2020 SC1-PM-15-2017 (n° 769643), and Validithi EIT Health Program. The founders had no role in the study design, data collection, analysis, and preparation of the manuscript.

\bibliographystyle{IEEEtran}
\bibliography{bibliography.bib}
\end{document}